\documentclass[conference]{IEEEtran}
\IEEEoverridecommandlockouts
\usepackage{amsmath,amssymb,amsfonts}
\usepackage{algorithmic}
\usepackage{graphicx}
\usepackage{textcomp}
\usepackage{xcolor}

\usepackage{CJKutf8}
\usepackage{CJK}

\usepackage{tabularx}
\usepackage{booktabs}
\usepackage{url}

\def\BibTeX{{\rm B\kern-.05em{\sc i\kern-.025em b}\kern-.08em
    T\kern-.1667em\lower.7ex\hbox{E}\kern-.125emX}}
\begin{document}

\title{Large Language Model Should Understand Pinyin \\for Chinese ASR Error Correction}

\author{Yuang Li, Xiaosong Qiao, Xiaofeng Zhao, Huan Zhao, Wei Tang, Min Zhang, Hao Yang\\ Huawei Translation Services Center, China\\\small{\{liyuang3, qiaoxiaosong, zhaoxiaofeng14, zhaohuan54, tangwei133, zhangmin186, yanghao30\}@huawei.com}}

% \address{$^{\star}$ Affiliation Number One \\
% $^{\dagger}$}Affiliation Number Two

% \and
% \and
% \and
% \and
% \and

\maketitle

\begin{abstract}

Large language models (LLMs) can enhance automatic speech recognition (ASR) systems through generative error correction (GEC). In this paper, we propose Pinyin-enhanced GEC (PY-GEC), which leverages Pinyin—the phonetic representation of Mandarin Chinese—as supplementary information to improve Chinese ASR error correction. Our approach only utilizes synthetic errors for training and employs the one-best hypothesis during inference. Additionally, we introduce a multitask training approach involving conversion tasks between Pinyin and text to align their feature spaces. Experiments on the Aishell-1 and the Common Voice datasets demonstrate that our approach consistently outperforms GEC with text-only input. More importantly, we provide intuitive explanations for the effectiveness of PY-GEC and multitask training from two aspects: 1) increased attention weight on Pinyin features; and 2) aligned feature space between Pinyin and text hidden states.

\end{abstract}
\section{Introduction}

End-to-end architectures~\cite{chorowski2015attention, graves2006connectionist, graves2012sequence} have been widely adopted in automatic speech recognition (ASR). However, several factors can lead to low-quality ASR outputs, such as environmental noise, speech overlaps, long-tail words, and speaker accents. Therefore, researchers have proposed various methods to correct ASR outputs~\cite{leng-etal-2021-fastcorrect-2, 2024Jin, 2022zheng, ma2023generative, 2023yang}. Among these, using large language models (LLMs) for generative error correction (GEC) has gained traction due to LLMs’ strong performance across diverse tasks such as text rewriting~\cite{Shu_2024}, grammar correction~\cite{Fan_2023}, and spoken language understanding~\cite{2024li,qwenaudio2023}. The LLM-based GEC involves directly feeding the LLM with the N-best hypotheses and prompting it to perform rerank and correction simultaneously~\cite{ma2023generative, 2023yang}. To further enhance GEC performance, audio features can be incorporated by training an adapter layer~\cite{radhakrishnan2023whispering}.

ASR errors, unlike typographical and grammatical errors, often involve misrecognizing one word as another due to similar pronunciation. Consequently, Chinese ASR error correction poses a challenge because there is no direct connection between the pronunciation and the written form of Chinese characters. To help the model grasp the semantic meaning and pronunciation of the Chinese transcriptions for accurate correction, Pinyin, which uses the Latin alphabet to represent phonetics, can be used as input. Previous research has explored methods such as direct Pinyin recognition from speech input followed by Pinyin to text conversion~\cite{2021yuan}, error recognition followed by Pinyin mask filling~\cite{2022zheng}, and the projection of text and Pinyin features to the same space using contrastive learning and a shared encoder~\cite{2024Jin}. We extend the use of Pinyin to the LLM and introduce Pinyin-enhanced generative error correction (PY-GER), which leverages Pinyin features and enhances LLM's comprehension of Pinyin through multitask training.

The most relevant study to our work is Pinyin Regularization~\cite{2024tang} which uses the Pinyin of the N-best hypotheses for ChatGPT~\cite{gpt2023gpt} and ChatGLM~\cite{zeng2023glm130b} to enhance Chinese ASR error correction. Our approach differs in several key aspects: we exclusively utilize the one-best hypothesis instead of the N-best hypotheses, employ pseudo ASR errors for model training rather than real ASR errors, and incorporate multitasking. Furthermore, we provide an extensive analysis to elucidate the rationale and mechanisms behind the effectiveness of Pinyin.

\begin{figure}[t]
  \centering
  \includegraphics[width=0.65\linewidth]{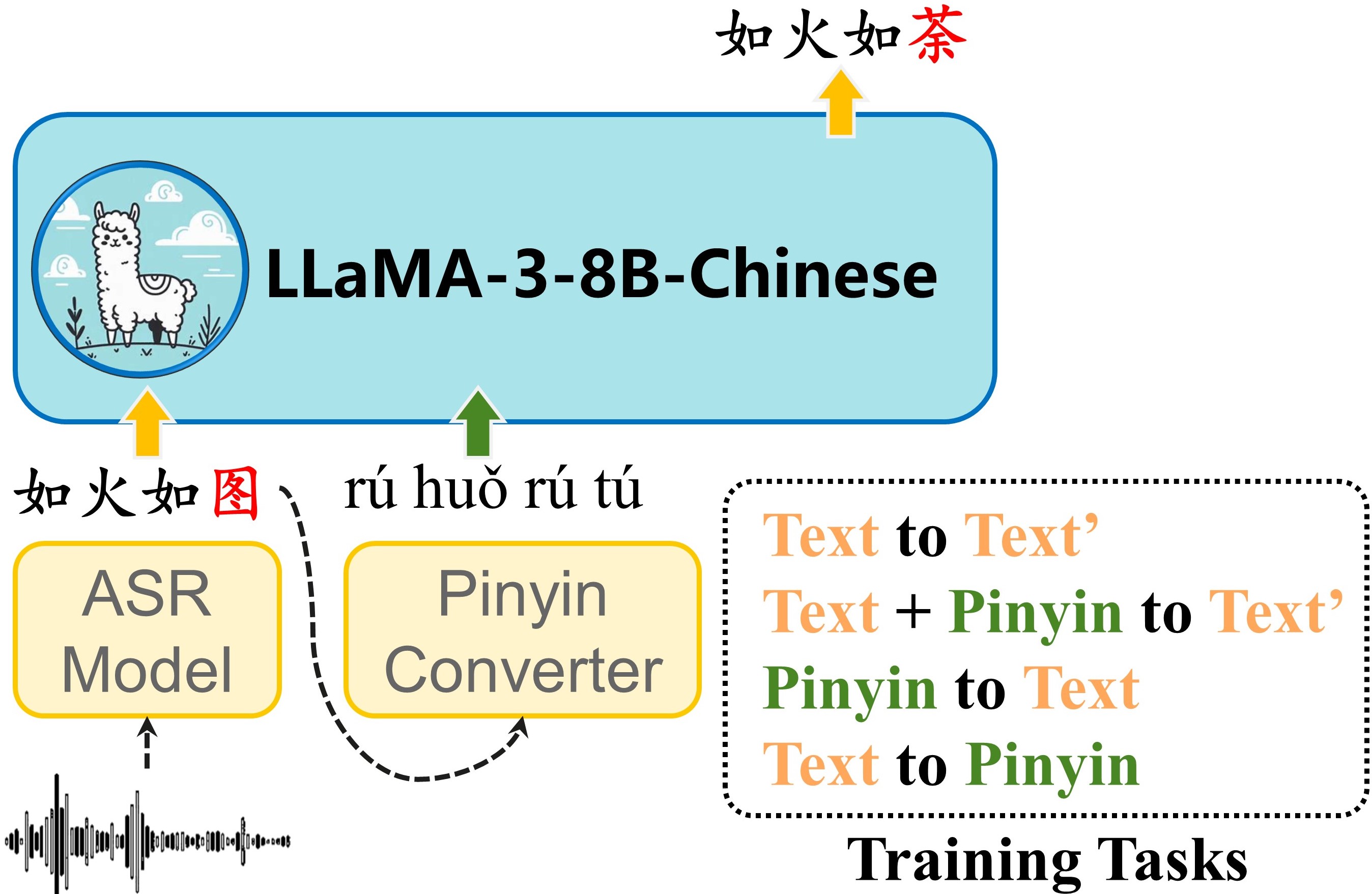}
  \caption{The flowchart for PY-GEC.}
  \label{fig:method}
\end{figure}

In this paper, we introduce PY-GEC with multitask training and carry out experiments on the Aishell-1~\cite{aishell} and Common Voice datasets~\cite{ardila-etal-2020-common} using the transcriptions generated from the Whisper-Small and the Whisper-Large-v2 models~\cite{radford2022robust}. Our findings demonstrate that incorporating Pinyin consistently improves the character error rates (CERs) and entity recalls. Furthermore, multitask training enhances overall performance and contributes to a relative CER reduction of 8.3\% and a relative entity recall improvement of 3.9\% on average compared to direct correction. Additionally, we explore combining multiple corrected results from the multitask-trained model to achieve further performance enhancements. Notably, all training is performed on a text-only synthetic dataset, which is created without access to ASR models or speech data. To demonstrate the efficacy of Pinyin, we calculate attention scores between the output and the input text, the output and the input Pinyin, as well as the output and itself. We reveal that the proposed method assigns the highest attention weight to Pinyin features. Additionally, we employ a straightforward yet effective downsampling technique to quantify and visualize the alignment between the hidden states of Pinyin and Text. Notably, our approach successfully projects Pinyin features into a feature space most similar to that of the text features.

% The paper proceeds as follows: Section 2 introduces PY-GEC methodologies and visualization techniques. Section 3 outlines the training and evaluation setups, while Section 4 presents the experimental results and findings. Finally, Section 5 concludes the paper.

\section{Methodology}

\subsection{PY-GEC and Multitask Training}

The flowchart for PY-GEC is depicted in Figure~\ref{fig:method}. The one-best transcription of the input speech signal serves as input to the LLM and is also converted to Pinyin, which acts as supplementary input. The LLM leverages both semantic and phonetic information to generate the corrected output.

To train the LLM for PY-GEC, we introduce multitask training with the following tasks: \textbf{1) Direct Correction}: The LLM predicts the corrected output based on the one-best hypothesis. \textbf{2) PY-GEC}:  The LLM predicts the corrected output by considering both the one-best hypothesis and its corresponding Pinyin representation. \textbf{3) Pinyin to text conversion}: The LLM converts Pinyin to its corresponding text. Additionally, we use the Pinyin associated with the hypothesis and convert it to the ground truth text, allowing the LLM to better understand erroneous Pinyin. \textbf{4) Text to Pinyin conversion}: The LLM converts text to its corresponding Pinyin representation.

The correction tasks promote the LLM's ability to recognize and correct ASR errors while the conversion tasks enhance the LLM's understanding of the alignment between text and Pinyin. The prompt for each task is provided in Table~\ref{table:prompt}.

\begin{CJK}{UTF8}{gbsn}
\begin{table}[t]
\caption{Prompts for GEC and multitask training.}
% \small

    \centering
    \begin{tabular}{p{1.8cm} | p{5.5cm}} % Adjust the width of the second column as per your requirement
    \toprule
    Direct & 请改正转录文本。 转录文本：[hypothesis]\\
    
    (translation) & Please correct the transcription. Transcription: [hypothesis]\\
    \midrule
    PY-GEC & 请根据转录文本的拼音，改正转录文本。（注意同音词的错误）转录文本：[hypothesis]\ 拼音：[Pinyin]\\
    (translation) & Please correct the transcription according to its Pinyin. (Note errors in homophones) Transcription: [hypothesis] Pinyin: [Pinyin]\\
    \midrule
    Pinyin to text & 请将拼音转化为文本。拼音：[Pinyin of reference or hypothesis]\\
    (translation) & Please convert pinyin to text. Pinyin: [Pinyin of reference or hypothesis]\\
    Text to pinyin & 请将文本转化为拼音。文本：[reference]\\
    (translation) & Please convert the text to pinyin. Text: [reference]\\
    \bottomrule
    \end{tabular}
    \label{table:prompt}
\end{table}
\end{CJK}

\subsection{Pseudo Dataset}
\label{sec:data}

Due to Chinese homophones, most errors in Chinese ASR are substitutions. In our pilot study, substitution errors can be 20 times more than deletion and insertion errors on the Aishell-1 dataset~\cite{aishell}. Consequently, when creating an ASR error correction dataset, we primarily focus on substitution errors. We preprocess the training set text by tokenizing sentences into words and filtering out high-frequency words, which are commonly recognized accurately by the ASR system. Subsequently, we randomly select a subset of sentences, and for each sentence, we choose words at random and replace characters based on a homophone dictionary~\footnote{\url{https://github.com/LiangsLi/ChineseHomophones}}.

% https://github.com/LiangsLi/ChineseHomophones
% 40% 0.1 to 0.4 ratio
% 去掉top 5000，剩下
% aishell 120098, common voice 16499, total 136597
% 80621 words
% 409791

% \begin{table}[t]
% \caption{The number of errors on Aishell-1 dataset.}
% \small

% \centering
% \begin{tabular}{c | c c c} 
% \toprule
% & \textbf{substitutions} & deletions & insertions \\
% \midrule
% Whisper-Small & \textbf{10436} & 464 & 795\\
% Whisper-Large & \textbf{5119} & 466 & 656\\
% \bottomrule
% \end{tabular}
% \label{table:errors}
% \end{table}

% \begin{algorithm}
% \caption{Pseudo Dataset Creation}
% \begin{algorithmic}[1]
% \REQUIRE text $t$, keywords $\mathbf{w}$, substitution probability $p_s$, substitution ratio $(s_l, s_u)$, confusion dictionary
% \STATE $\mathbf{w}_t \gets segment(t) $
% \STATE $candidates \gets \mathbf{w}_t \cap \mathbf{w}$
% \FOR{$c \in candidates$}
%     \IF{$\mathcal{U}(0,1) < p_s$}
%     \STATE $n \gets \lceil len(t)\times\mathcal{U}(s_l, s_u) \rceil$
%     \STATE $\hat{c} \gets$ Randomly replace $n$ characters in $c$ based on the confusion dictionary.
%     \STATE $\hat{t} \gets$ Replace $c$ with $\hat{c}$ in $t$.
%     \ENDIF
% \ENDFOR
% \RETURN $t, \ \hat{t}$
% \end{algorithmic}
% \label{alg:aug}
% \end{algorithm}

\subsection{Ensemble}

After multitask training, the LLM can perform GEC using information from various sources, including text-only and Pinyin-only data, as well as a combination of text and Pinyin. Consequently, we can ensemble multiple results generated from these diverse information sources. Specifically, we employ three methods: ROVER~\cite{rover}, LLM-rerank~\cite{10389732}, and a novel Pinyin-rerank method. In the Pinyin-rerank method, we convert both the predictions and the input text to Pinyin and then compute the CER between the Pinyin of each prediction and the rest of the predictions, as well as the input (Equation~\ref{eq:2}). We select the result with the lowest score. This method assumes that the corrected text’s Pinyin should be similar to the Pinyin of other predictions and the input, thus preventing hallucinations.

% \begin{align}
% \label{eq:1}
% \text{score}_{LLM} = \sum_{i=1}^{N} \log P(w_i | \mathbf{w}_{<i})
% \end{align}

\begin{align}
\label{eq:2}
\text{score}_{Pinyin} = \sum_{j=1}^{M} CER(Pinyin(\mathbf{w}_{j}), Pinyin(\mathbf{w}))
\end{align}

% LLM rerank logit公式
% Pinyin rerank公式

\subsection{Analysis}
\label{sec:align}

To interpret the effectiveness of Pinyin features, we compute the sum of attention scores across layers and attention heads. The attention score is computed between 1) the output and the input text; 2) the output and the input Pinyin; and 3) the output and itself. These attention scores can be regarded as the importance of different components, including the context from ASR transcription, the phonetic information, and the unidirectional context of the output.

% \begin{align}
% \label{eq:3}
% \text{Attention}(x, y) = \sum_{l=1}^{M}\sum_{h=1}^{N} \sum \frac{{\mathbf{Q}^{(l, h)}_x \cdot \mathbf{K}_y^{(l, h)T}}}{{\sqrt{d_k}}}
% \end{align}

% \noindent where M is the number of layer and N is the number of attention heads.

% Feature Space
To explore the relationship between the feature spaces of text and Pinyin, we compress their hidden states into feature vectors (Equation~\ref{eq:4}~\ref{eq:5}). For text features, we employ straightforward average pooling. However, since Pinyin features are typically longer than text features, simple average pooling fails to yield a representative vector. To address this, we downsample the Pinyin features to match the length of the text features by selecting the hidden states with the highest cosine similarities to the text feature vector and then performing average pooling. Finally, we quantify the text-Pinyin alignment using cosine similarity between their feature vectors and provide visualizations with principal component analysis (PCA).

\begin{align}
\label{eq:4}
\mathbf{v}_{text} &= \frac{1}{T} \sum \mathbf{H}_t \\
\label{eq:5}
\mathbf{v}_{Pinyin} &= \frac{1}{T} \sum downsample(\mathbf{H}_p, \mathbf{v}_{text})
\end{align}

\noindent where $T$ is the length of text hidden states $\mathbf{H}_t$. $\mathbf{H}_p$ is the Pinyin hidden states.

\begin{table*}[t]
\caption{The performance of error correction is measured by \textbf{CER and Entity Recall}. ‘Re-transcribe’ means that the LLM only sees the Pinyin. ‘Ensemble’ refers to merging or reranking the three results from the multitask-trained model.}
\small
\centering
\begin{tabular}{c | c c | c c | c} 
\toprule
& \multicolumn{2}{c|}{Aishell-1} & \multicolumn{2}{c|}{Common Voice} &  \\
& Whisper-Small & Whisper-Large & Whisper-Small & Whisper-Large & Average \\
\bottomrule
No GEC & 11.16 / 62.60 & 5.96 / 74.92 & 22.21 / 56.51 & 14.32 / 69.75 & 13.43 / 65.95\\
Direct & 10.03 / 67.85 & 5.78 / 76.98 & 18.32 / 63.02 & 11.80 / 72.92 & 11.48 / 70.19 \\
PY-GEC & 9.61 / 68.46 & 5.64 / 77.42 & 17.67 / 64.08 & 11.48 / 73.30 & 11.10 / 70.82\\
\midrule
Multitask + Direct & 9.61 / 70.76 & 5.70 / 78.34 & 18.01 / 64.02 & 11.77 / 72.81 & 11.27 / 71.48\\
Multitask + Re-transcribe & 9.66 / 71.12 & 7.20 / 76.57 & 19.28 / 62.82 & 14.46 / 68.91 & 12.65 / 69.86\\
\textbf{Multitask + PY-GEC} & \textbf{8.63} / \textbf{72.97} & \textbf{5.39} / \textbf{78.96} & \textbf{16.84} / \textbf{65.84} & \textbf{11.27} / \textbf{73.94} & \textbf{10.53} / \textbf{72.93} \\
\midrule
Ensemble (ROVER) & 8.93 / 72.94 & 5.60 / 78.87 & 17.67 / 64.87 & 11.76 / 73.33 & 10.99 / 72.50\\
Ensemble (Pinyin-Rerank) & 8.46 / 73.06 & \textbf{5.31} / 79.02 & \textbf{16.61} / 66.36 & \textbf{10.99} / 74.42 & \textbf{10.34} / 73.22\\
Ensemble (LLM-Rerank) & \textbf{8.36} / \textbf{74.21} & 5.41 / \textbf{80.46} & 16.72 / \textbf{66.81} & 11.26 / \textbf{74.51} & 10.44 / \textbf{74.00}\\
\bottomrule
\end{tabular}

\label{table:asr1}
\end{table*}

\section{Experiments}

\subsection{Setups}

We utilize two datasets: Aishell-1~\cite{aishell} and Common Voice~\cite{ardila-etal-2020-common}. Our training set is derived from the text data in their training sets. We extract 80,621 words, filter out the top 5,000 most frequent words, and introduce errors with a 40\% probability, as detailed in Section~\ref{sec:data}. For training, we synthesize a total of 136,597 samples. As for the test set, Aishell-1 and Common Voice contain 7,176 and 8,273 samples, respectively. During the training phase, we perform fine-tuning on the LLaMA-3-8B-Chinese model~\footnote{\url{https://huggingface.co/shenzhi-wang/Llama3-8B-Chinese-Chat}} for a single epoch, using a learning rate of 1e-4, a batch size of 16, and a LoRA rank of 32~\cite{hu2022lora, zheng2024llamafactory}. For evaluation, we employ greedy decoding and select the one-best hypothesis generated by Whisper-Small and Whisper-Large-v2~\cite{radford2022robust}, which are advanced ASR models trained on a massive speech corpus, as the input. To evaluate performance, we employ two metrics: character error rate (CER) and entity recall. CER provides a measure of overall ASR performance, while entity recall assesses the ability to recognize keywords. For the Aishell dataset, we utilize entity labels from the Aishell-NER dataset~\cite{Chen_2022}. For the Common Voice dataset, we rely on predicted entity labels generated by a NER model. Additionally, we analyze the percentage of good and bad cases where the CERs are reduced and increased by the LLM respectively.

% dataset matric / commonvoice 提取实体 /CER RECALL GOOD BAD CASE
% LLAMA Chinese介绍，whisper介绍 用的small和large-v2
% 1 epoch / 1e-4 / bs 1 * 8 * 2 / rank 32
% https://huggingface.co/shenzhi-wang/Llama3-8B-Chinese-Chat
% https://github.com/hiyouga/LLaMA-Factory

% https://github.com/mozillazg/python-pinyin/tree/master

% https://github.com/LiangsLi/ChineseHomophones
% 40% 0.1 to 0.4 ratio
% 去掉top 5000，剩下
% aishell 120098, common voice 16499, total 136597
% 80621 words
% 409791

\subsection{Results for PY-GEC}

\begin{figure}[t]
  \centering
  \includegraphics[width=0.85\linewidth]{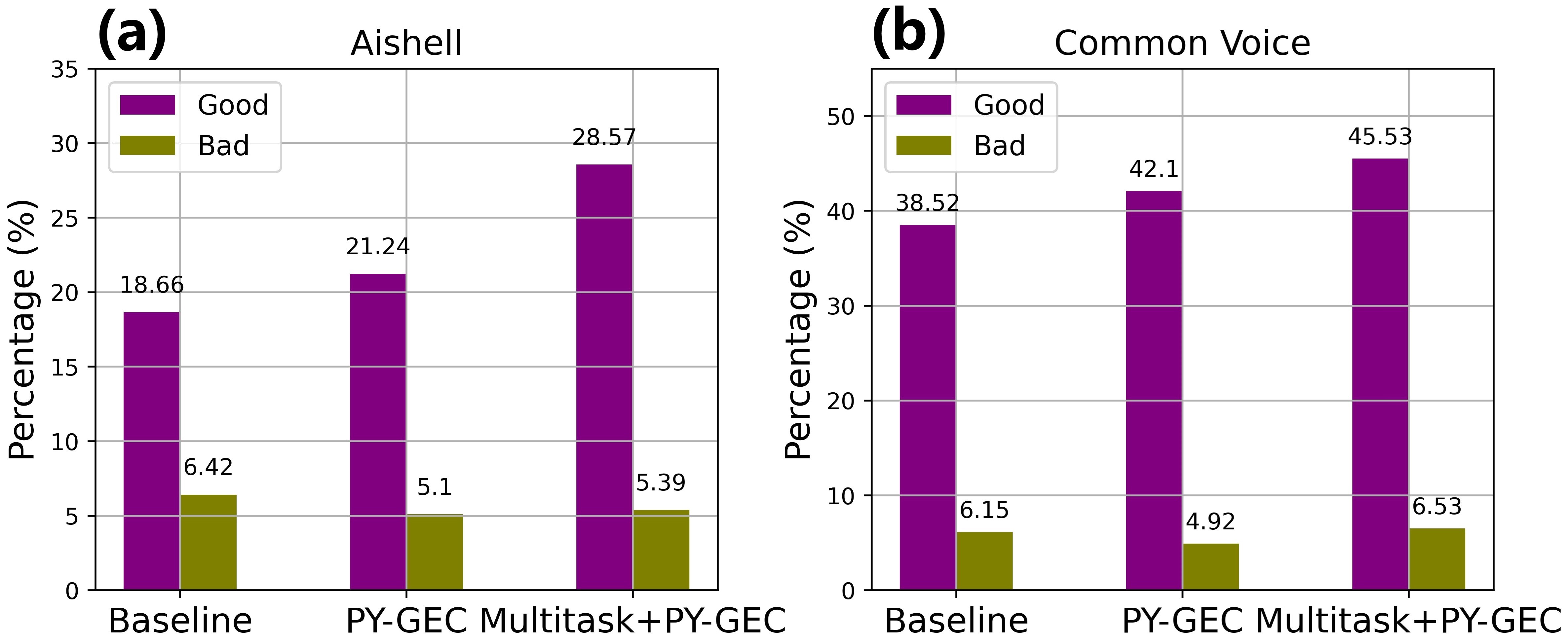}
  \caption{The percentages of good and bad cases. The ASR transcriptions are generated by Whisper-Small on Aishell-1 and Common Voice datasets.}
  \label{fig:bar}
\end{figure}

In Table~\ref{table:asr1}, we observe consistent improvements in CERs and entity recalls across all test sets using LLM-based correction methods. Specifically, direct correction enhances the average CER and entity recall from 13.43\% and 64.95\% to 11.48\% and 70.19\%, respectively. Furthermore, the PY-GEC method achieves even better performance, with a CER of 11.10\% and an entity recall of 70.82\%. Notably, when using multitask training and PY-GEC, we achieve the lowest CER and the highest entity recall across all ASR models and test sets. The average CER and entity recall reach 10.53\% and 72.93\%, respectively, proving the effectiveness of our proposed approach. Analyzing Figure~\ref{fig:bar}, we find that on the Aishell dataset, multitask training and PY-GEC significantly improve the percentage of good cases from 18.66\% to 28.57\%, while reducing bad cases from 6.42\% to 5.39\%. On the Common Voice dataset, the percentage of good cases increases from 38.52\% to 45.53\%, while bad cases remain relatively stable. After multitask training, the LLM can perform direct correction with text-only input and re-transcription from Pinyin-only input. Surprisingly, when using multitask training, direct correction performs comparably to PY-GEC and outperforms direct correction without multitask training. This suggests that multitask training enhances the LLM’s internal understanding of input text's pronunciation, improving its ability to recognize and correct ASR errors. Furthermore, retranscription can improve the ASR performance of the Whisper-Small model but not the Whisper-Large model. Ensembling results from the multitask-trained LLM is generally effective. However, the traditional sequence merging method, ROVER, does not enhance the CER or entity recall. In contrast, the Pinyin-Rerank method consistently improves both CER and entity recall across all setups. Although LLM-Rerank achieves the highest entity recall, its CER is higher than that of Pinyin-Rerank and requires more computational resources.

\subsection{Attention Analysis}

\begin{figure}[t]
  \centering
  \includegraphics[width=0.5\linewidth]{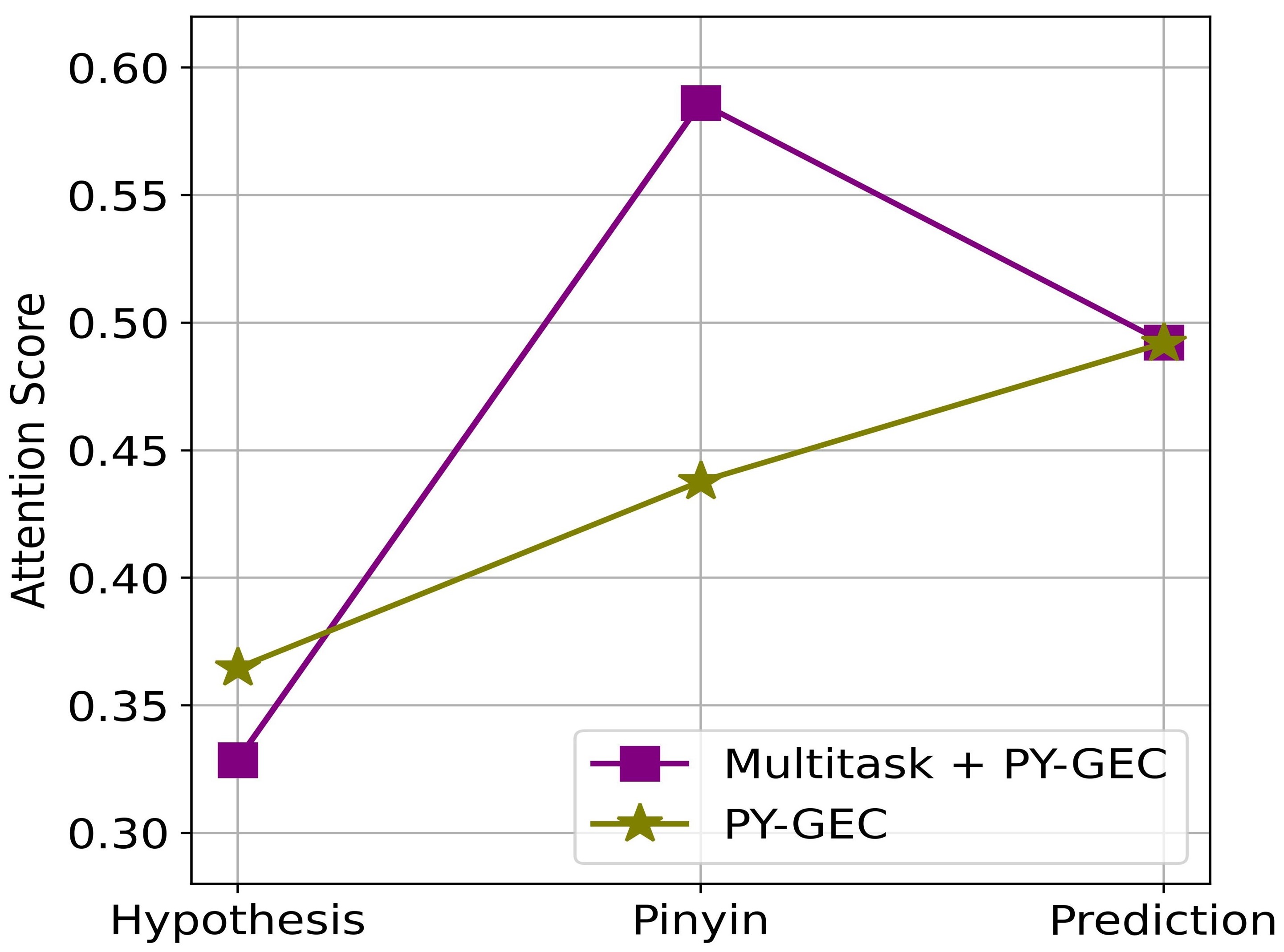}
  \caption{The attention scores correspond to each input component.}
  \label{fig:att}
\end{figure}

The attention scores depicted in Figure~\ref{fig:att} shed light on the significance of each input component in the error correction process. Notably, for the naive PY-GEC approach, Pinyin exerts a more substantial influence than the input hypothesis. However, it is the prediction that receives the highest attention score. This observation suggests that the GEC process predominantly relies on the context provided by nearby output tokens. Multitask training enhances the importance of Pinyin features, as indicated by the highest attention score, demonstrating that the LLM better comprehends Pinyin features.

\begin{figure}[t]
  \centering
  \includegraphics[width=0.85\linewidth]{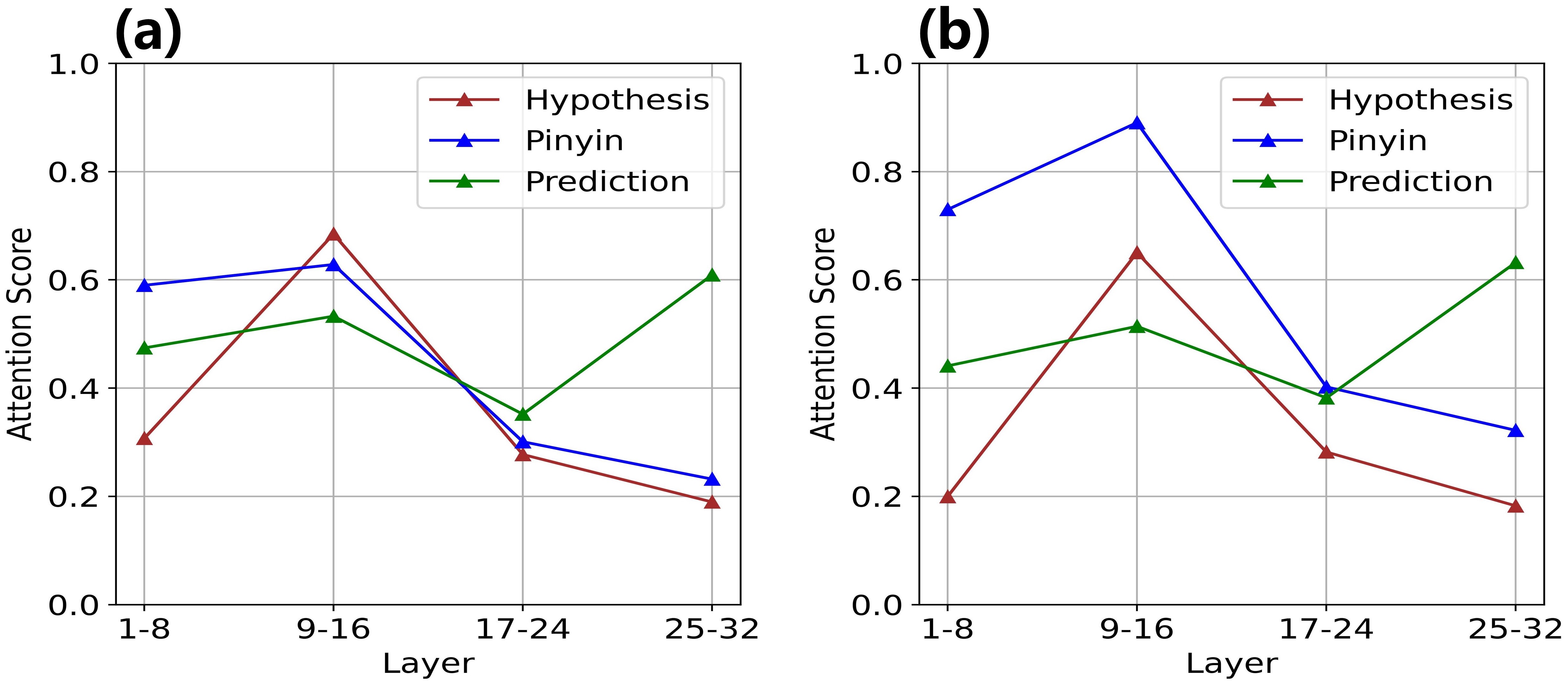}
  \caption{The layer-wise attention scores correspond to each input component. (a) PY-GEC; (b) Multitask + PY-GEC.}
  \label{fig:att_layer}
\end{figure}

Figure~\ref{fig:att_layer} illustrates the layer-wise attention scores. For PY-GEC, we observe that the importance of each component remains similar across the initial 24 layers. However, as we delve deeper into the network, the predicted tokens receive increased attention. In the multitask training scenario, Pinyin features consistently exhibit higher attention scores across all layers compared to the features of the hypothesis, while predicted tokens continue to play a crucial role at deeper layers

\subsection{Feature Space Analysis}

In Table~\ref{table:sim}, we evaluate the alignment between text and Pinyin, as outlined in section~\ref{sec:align}. Initially, without fine-tuning, the LLaMA-3-8B-Chinese shows poor alignment with a low cosine similarity of 0.26. Fine-tuning with PY-GEC or multitask training can both significantly boost the alignment with cosine similarity improved to 0.74 and 0.82 respectively. These results also verify the benefits of incorporating conversion tasks to enhance text-Pinyin alignment. Unexpectedly, even the model fine-tuned with direct correction demonstrates better text-Pinyin alignment. Despite not having been exposed to Pinyin features during training, this model likely learns Chinese character pronunciation from the ASR correction task, closing the gap between text and Pinyin. This further emphasizes the importance of promoting phonetic representation understanding within large language models for better correction performance.

\begin{figure}[t]
  \centering
  \includegraphics[width=0.75\linewidth]{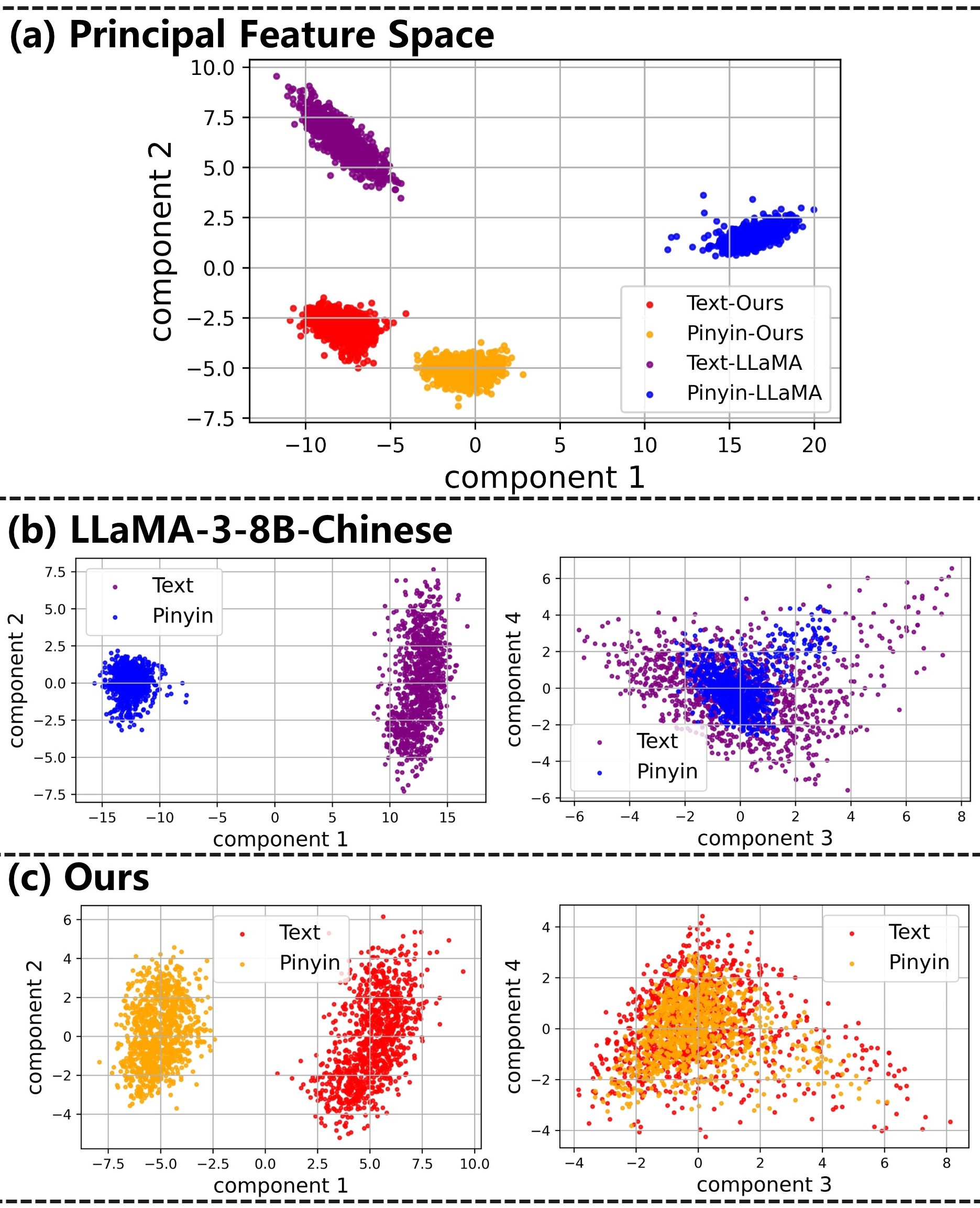}
  \caption{PCA analysis for the last hidden states that correspond to Text and Pinyin. The hidden states are extracted from the original LLaMA-3-8B-Chinese model and our fine-tuned multitask model.}
  \label{fig:pca}
\end{figure}

Figure~\ref{fig:pca} illustrates the feature space of text and Pinyin. In Figure~\ref{fig:pca} (a), we observe that our approach (multitask + PY-GEC) brings the feature space between text and Pinyin much closer than the original LLM. However, text and Pinyin still occupy distinct feature spaces, indicating that the LLM perceives semantic and phonetic information differently. Further analysis of higher-dimensional feature spaces (Figure~\ref{fig:pca} (b, c)) reveals that the original LLM places text samples in sparser regions compared to Pinyin samples. In contrast, our fine-tuned model exhibits similar spatial distributions for text and Pinyin, with clusters showing comparable shapes except for the first principal component.

\begin{table}[t]
\caption{The cosine similarity between the text and the Pinyin vectors.}
\small
\centering
\begin{tabular}{c | c} 
\toprule
& cosine similarity \\
\midrule
LLaMA-3-8B-Chinese & 0.26\\
Direct & 0.45 \\
PY-GEC & 0.74\\
\textbf{Multitask + PY-GEC} & \textbf{0.82}\\
\bottomrule
\end{tabular}

\label{table:sim}
\end{table}

\section{Conclusion}

In this study, we introduce PY-GEC, a novel Chinese ASR error correction method that leverages Pinyin features and employs multitask training for the LLM. Our emphasis lies in promoting LLM's understanding of the alignment between text and Pinyin features. We not only show the superiority of our approach but conduct a thorough analysis of attention scores and feature spaces, to elucidate the importance of Pinyin and text-Pinyin alignment. For future research, we aim to extend our experiments to larger-scale LLMs and multi-modal LLMs.

\clearpage

\bibliographystyle{IEEEbib}
\bibliography{refs}

\end{document}